\documentclass[letterpaper]{article} 
\usepackage{aaai24}  
\usepackage{times}  
\usepackage{helvet}  
\usepackage{courier}  
\usepackage[hyphens]{url}  
\usepackage{graphicx} 
\usepackage{natbib}  
\usepackage{caption} 
\usepackage{algorithm}
\usepackage{algorithmic}
\usepackage{mathtools}
\usepackage{graphicx}
\usepackage{multirow}
\usepackage{booktabs}
\usepackage{pifont}
\usepackage{tikz}
\usepackage{amsthm}

\newtheorem{prop}{Proposition}
\usepackage{newfloat}
\usepackage{listings}
\DeclareCaptionStyle{ruled}{labelfont=normalfont,labelsep=colon,strut=off} 
\floatstyle{ruled}
\newfloat{listing}{tb}{lst}{}
\floatname{listing}{Listing}
\title{Cross-Modal and Uni-Modal Soft-Label Alignment for Image-Text Retrieval}
\author {
    Hailang Huang\textsuperscript{\rm 1},
    Zhijie Nie\textsuperscript{\rm 1,2},
    Ziqiao Wang\textsuperscript{\rm 3},
    Ziyu Shang\textsuperscript{\rm 4}
}
\affiliations {
    \textsuperscript{\rm 1}SKLSDE, School of Computer Science and Engineering, Beihang University, Beijing, China\\
    \textsuperscript{\rm 2}Shen Yuan Honors College, Beihang University, Beijing, China \\
    \textsuperscript{\rm 3}School of Electrical Engineering and Computer Science, University of Ottawa, Ottawa, Canada \\
    \textsuperscript{\rm 4}School of Computer Science and Engineering, Southeast University, Nanjing, China \\
    \{huanghl, niezj\}@act.buaa.edu.cn, zwang286@uottawa.ca, ziyus1999@seu.edu.cn
}

\usepackage{bibentry}

\begin{document}

\maketitle

\begin{abstract}
Current image-text retrieval methods have demonstrated impressive performance in recent years. However, they still face two problems: the inter-modal matching missing problem and the intra-modal semantic loss problem. These problems can significantly affect the accuracy of image-text retrieval. To address these challenges, we propose a novel method called Cross-modal and Uni-modal Soft-label Alignment (CUSA). Our method leverages the power of uni-modal pre-trained models to provide soft-label supervision signals for the image-text retrieval model. Additionally, we introduce two alignment techniques, Cross-modal Soft-label Alignment (CSA) and Uni-modal Soft-label Alignment (USA), to overcome false negatives and enhance similarity recognition between uni-modal samples. Our method is designed to be plug-and-play, meaning it can be easily applied to existing image-text retrieval models without changing their original architectures. Extensive experiments on various image-text retrieval models and datasets, we demonstrate that our method can consistently improve the performance of image-text retrieval and achieve new state-of-the-art results. Furthermore, our method can also boost the uni-modal retrieval performance of image-text retrieval models, enabling it to achieve universal retrieval. The code and supplementary files can be found at \url{https://github.com/lerogo/aaai24_itr_cusa}.

\end{abstract}

\section{Introduction}
Image-Text Retrieval (ITR) retrieves relevant samples from one modality based on a query in another modality. It involves two sub-tasks, one is image-to-text retrieval, which requires finding the most relevant caption in the text gallery for an input image, and the other one is text-to-image retrieval, which requires finding the most relevant image in the image gallery for an input query text. Most existing ITR methods \cite{clip_radford2021learning, sgraf_diao2021similarity, x2vlm_zeng2023x2vlm} adopt contrastive learning techniques, treating one sample as an anchor and the corresponding sample in the other modality as a positive sample, while the uncorrelated samples are considered negatives. These methods aim to maximize the similarity between anchor and positive samples and minimize the similarity between anchor and negative samples for cross-modal retrieval. Although these ITR methods have achieved impressive performance, they have two limitations: \textbf{the inter-modal matching missing problem} and \textbf{the intra-modal semantic loss problem}. 
\begin{figure}[t]
\centering
\includegraphics[width=\linewidth]{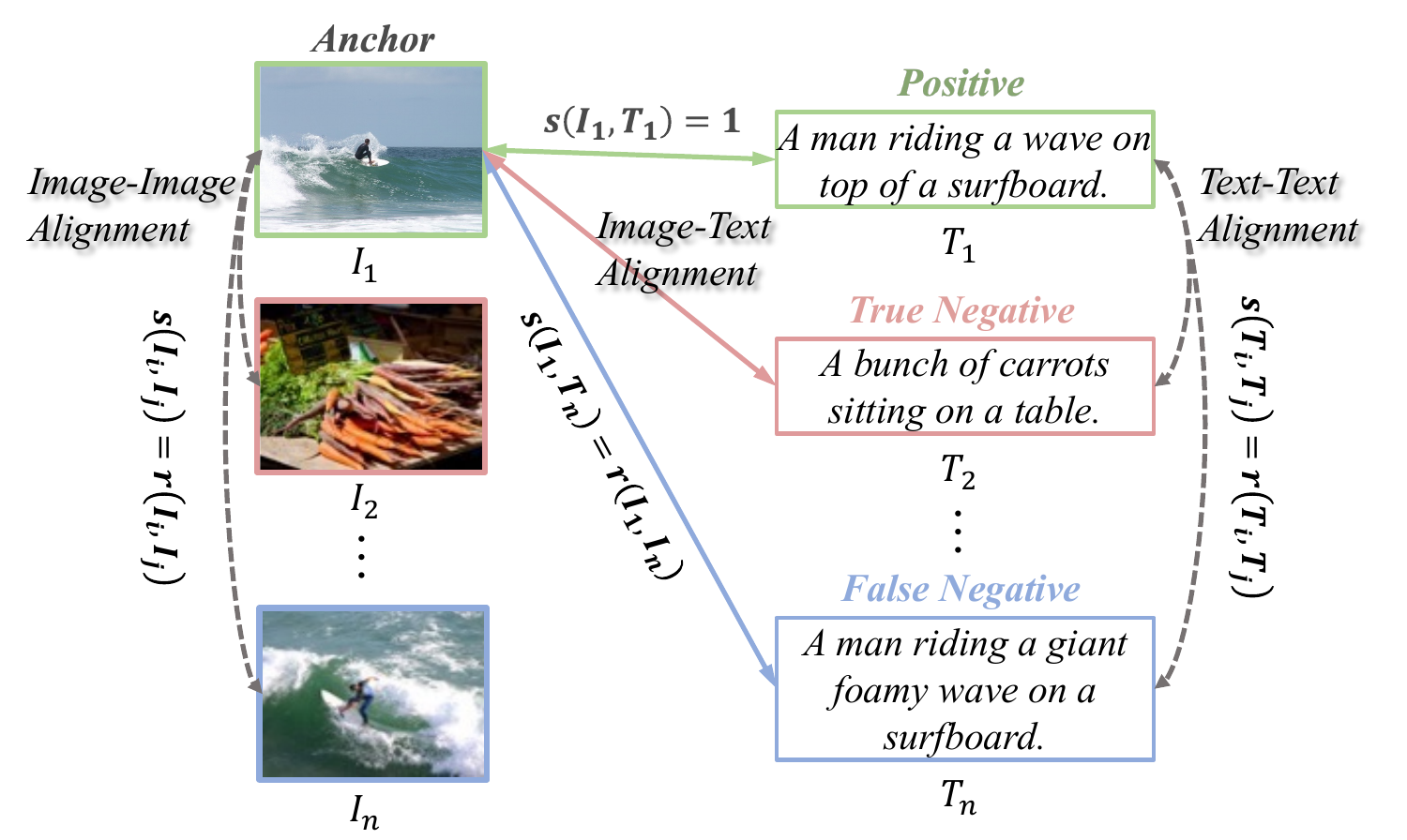}
\caption{Illustration of our approach. We use soft-labels $r(\cdot, \cdot)$ generated by uni-modal teacher models as a supervisory signal to guide cross-modal alignment and uni-modal alignment for image-text retrieval models.}
\label{fig:method_lite}
\end{figure}

The inter-modal matching missing problem refers to the situation where, during model training, samples that should be matched are mistakenly treated as unmatched due to contrastive learning techniques and random sampling, resulting in a decrease in model performance. As shown in Figure \ref{fig:method_lite}, in a batch containing $n$ image-text pairs, if the image $I_1$ is considered as the anchor, the traditional contrast learning used in most ITR methods treats the text $T_1$ as the positive sample matched with it (hard-labels), and all other texts $T_{j,j \neq 1}$ as negative samples, but text $T_n$ is matched with image $I_1$. Samples like $T_n$ are called false negative samples, and the noisy signals caused by such false negative samples will weaken the performance of the ITR model. All of the image-text pairs in Figure \ref{fig:method_lite} are sampled from the widely used MSCOCO \cite{lin2014microsoft} dataset, demonstrating the existence of the inter-modal matching missing problem. This problem has also attracted the attention of other researchers~\cite{cc_parekh2021crisscrossed,false_negative_Chun_2021_CVPR,chun2022eccv_caption}.~\cite{cc_parekh2021crisscrossed} published the first dataset CrissCrossed Caption (CxC) with soft-labels to correct false negative samples, but this dataset only focuses on scoring the similarity between texts, resulting in many missing positives in the text-to-image relationship. Therefore, \citet{chun2022eccv_caption} released a machine-and-human-verified dataset ECCV Caption, which includes manually verified corrections of false negatives, and pointed out that false negatives hinder model evaluation. Additionally, they recommended using the informative ranking-based metric mAP@R to evaluate model performance. 

The intra-modal semantic loss problem refers to the insufficient capability of current ITR models to recognize similar input samples. The reason for this problem is that most ITR methods solely prioritize optimizing the similarity between two modalities, disregarding the relationships within each modality. As illustrated in Figure \ref{fig:method_lite}, most ITR methods only focus on aligning image-text pairs and overlook the operations of image-image alignment and text-text alignment. TCL~\cite{tcl_yang2022vision} tries to introduce self-supervised contrastive learning~\cite{gao2022simcse, pmlr_simclr} methods on uni-modal to obtain better joint multi-modal features, and it implicitly performs uni-modal alignment at the same time, but our experiments show that it is difficult to achieve effective uni-modal alignment through data augmentation alone. We mathematically prove that solely emphasizing cross-modal alignment hinders the ability of the model to recognize similar input samples, thereby weakening the performance of image-text retrieval in cases where the model is confronted with unseen samples during training but similar to certain samples in the training set. Although there have been some methods~\cite{tcl_yang2022vision,li2022imagetext_bcls,li_zheng_integrating_lg,li2023integrating_pairwise} attempting to solve the above problems and achieve certain effects, they have only separately addressed one of the two problems without considering the correlation between them. 

To address the above two challenges, we propose a novel and comprehensive framework for image-text retrieval, called Cross-modal and Uni-modal Soft-label Alignment (CUSA). As shown in Figure \ref{fig:method_lite}, our method leverages uni-modal pre-training models to provide soft-label supervision signals for the ITR model. Compared to hard-labels, soft-labels can capture more fine-grained and nuanced semantic information across and within modalities. Our method uses two alignment techniques, Cross-modal Soft-label Alignment (CSA) and Uni-modal Soft-label Alignment (USA). The CSA method as a regularization term to guide the cross-modal alignment of the ITR model through soft-labels. With this approach, the model can learn not only from binary labels but also from continuous labels that reflect the semantic relatedness between images and texts. The USA method uses soft-labels to guide the uni-modal alignment of the ITR model. As a result, the model can better recognize similar samples within each modality and distinguish dissimilar ones. Our method is plug-and-play and can be easily applied to existing ITR models without changing their original architectures. We conduct extensive experiments on various ITR models and datasets and demonstrate that our method can consistently improve the performance of image-text retrieval and achieve new state-of-the-art results. Moreover, our method can also boost the uni-modal retrieval performance of the ITR model, enabling it to achieve universal retrieval. Our main contributions are summarized as follows:
\begin{itemize}
    \item We mathematically prove that solely emphasizing cross-modal alignment hinders the ability of the ITR model to recognize similar input samples, thereby weakening the performance of image-text retrieval.
    \item We introduce two alignment techniques, CSA and USA, that use soft-labels as supervision signals to guide the cross-modal and uni-modal alignment of the ITR model.
    \item We conduct extensive experiments on various ITR models and datasets, and show that our method can consistently improve the performance of image-text retrieval and achieve new state-of-the-art results.
\end{itemize}

\section{Related Work}
\paragraph{Image-Text Retrieval} Image-Text Retrieval (ITR) is a typical cross-modal task, whose main challenge is to learn a shared representation of images and texts and accurately measure their similarity. Existing models can be classified into three categories based on their architecture: (1) dual-encoder \cite{clip_radford2021learning, align_jia2021scaling}; (2) fusion-encoder \cite{scan_lee2018stacked,sgraf_diao2021similarity,vse00_chen2021learning,vsrnpp_li2022image,naaf_zhang2022negative}, and (3) dual-encoder + fusion-encoder \cite{albef_li2021align,blip_li2022blip,wang2022omnivl,x2vlm_zeng2023x2vlm}. Among them, dual-encoder models usually contain a text encoder and an image encoder, producing representations to measure the similarities between images and texts. Benefiting the simple calculation method, the dual-encoders model usually has a fast retrieval speed. However, due to the lack of interaction between images and texts, these models often have lower performance compared to fusion-encoder models. The models with both dual-encoder and fusion-encoder achieve a certain balance between performance and efficiency. Therefore, most recent works in ITR follow the dual-encoder or the dual-encoder+fusion-encoder architecture to ensure high retrieval efficiency. Recent works \cite{clip_radford2021learning,wang2022omnivl,x2vlm_zeng2023x2vlm} introduce self-supervised contrastive learning to align different modalities in the models with these two architectures. However, these models use hard-labels as supervised signals for training and only align images and texts annotated in the dataset, ignoring potential semantic similarities between different image-text pairs. Our method leverages the external knowledge provided by pre-trained uni-modal models, and it can be easily applied to existing ITR methods, thereby partially compensating for this limitation.

\paragraph{Alignmemt with Soft-label} Soft-label usually are used to alleviate the strict constraints imposed by noisy hard-labels and avoid excessive confidence in incorrect predictions by the model, which has been proven effective in various tasks. In the methods based on knowledge distillation, the logit produced by the teacher model can be regarded as the soft-label, which guides the learning of the student model. In the task of image-text retrieval, \citet{albef_li2021align} and \citet{gao2023softclip} use self-distillation~\cite{He_2020_CVPR_moco} to reduce the adverse effects of noisy image-text pairs. The core idea is to let the student model act as its teacher, and as training progresses, the student model dynamically evolves into its teacher. Besides, some methods \cite{li2022imagetext_bcls, li_zheng_integrating_lg, li2023integrating_pairwise} used external pre-trained language models to provide additional knowledge to overcome false negatives in contrastive learning. However, existing methods only use soft-labels for supervision in inter-modal, while our method is not only in it to align in inter-modal but also in intra-modal.

\section{Method}
\subsection{Preliminaries}
\begin{figure*}[t]
\centering
\includegraphics[width=.97\textwidth]{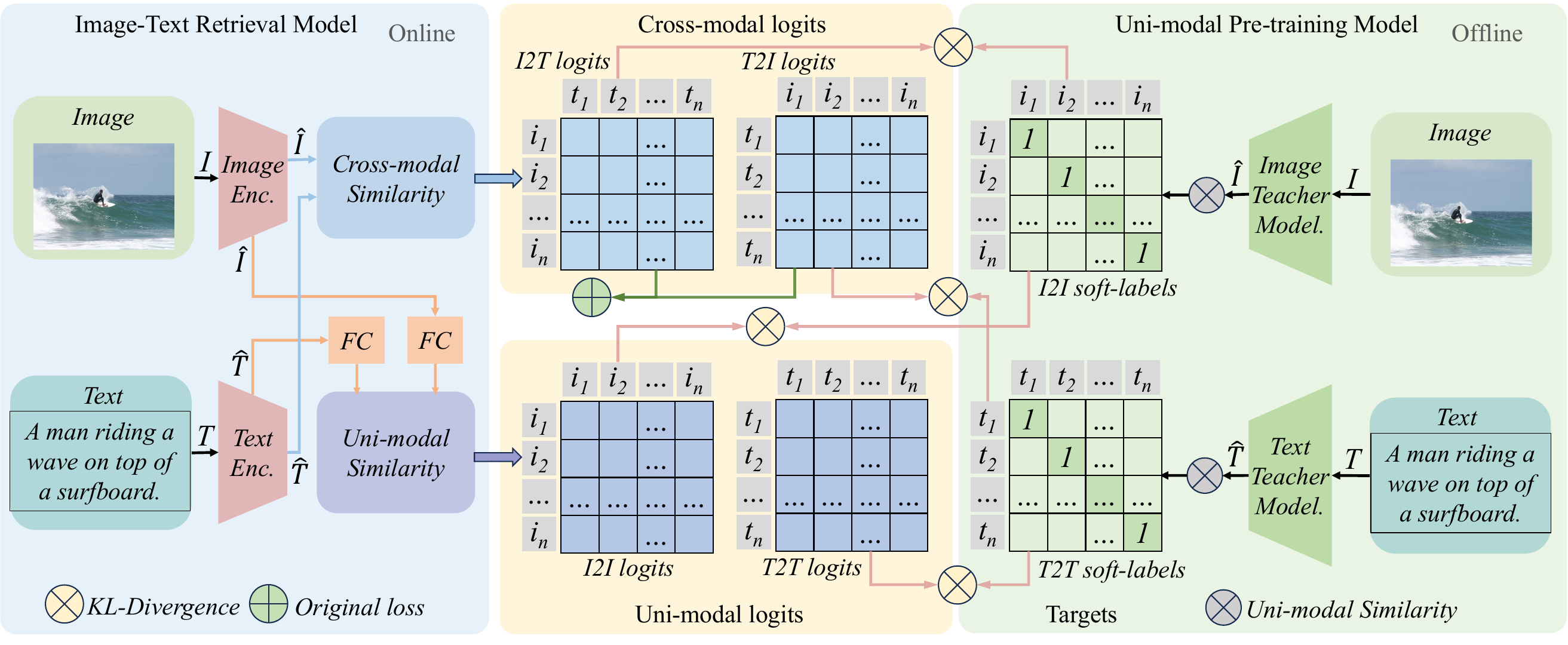}
\caption{Illustration of our proposed CUSA. It involves an ITR model used for training and a non-training uni-modal teacher model that provides soft-label supervision signals. The CSA method optimizes cross-modal logits, while the USA method optimizes uni-modal logits.}
\label{fig:method_all}
\end{figure*}

Given a dataset of image-text pair $\{(I_i, T_i)\}_{i=1}^N$, where $(I_i, T_i)$ represents the paired relationship between the image $I_i$ and the sentence $T_i$.  There is an image encoder that maps each $I_i$ to the normalized representation $\hat{I}_i$  and a text encoder that converts $T_i$ to the normalized representation $\hat{T}_i$.

The methods based on contrastive learning use InfoNCE loss \cite{infonce_oord2019representation} to align these image-text pairs. Specifically, multiple image-text pairs are sampled from the dataset and formed into a batch. During training, the paired image and text are ``pulled close'' while the unpaired ones in the batch are ``pushed away'' in the high-dimension space. We denote the cosine similarity of $\hat{I}_i$ and $\hat{T}_j$ as $s_{ij}^{\text{i2t}}$ obtained by the learnable encoders, then the probability that $I_i$ pairs with $T_j$ can be calculated by

\begin{equation}
\label{equ:sims_itc_i2t}
    Q_{ij}^{\text{i2t}}=\frac{\exp \left(s_{ij}^{\text{i2t}} / \tau\right)}{\sum_{k=1}^N \exp \left(s_{ik}^{\text{i2t}} / \tau\right)},
\end{equation}
where $N$ is the batch size and $\tau$ is a learnable temperature parameter. Similarly, we can denote the cosine similarity of $T_i$ and $I_j$ as $s_{ij}^{\text{t2i}}$ and calculate the probability that $T_i$ pairs with $I_j$ by
\begin{equation}
\label{equ:sims_itc_t2i}
    Q_{ij}^{\text{t2i}}=\frac{\exp \left(s_{ij}^{\text{t2i}} / \tau\right)}{\sum_{k=1}^N \exp \left(s_{ik}^{\text{t2i}} / \tau\right)}.
\end{equation}
Then we can obtain the discrete probability distribution $Q_{i}^{\text{i2t}}=(Q_{i1}^{\text{i2t}},...Q_{iN}^{\text{i2t}})$ for each image $I_i$ and $Q_{i}^{\text{t2i}}=(Q_{i1}^{\text{t2i}},...Q_{iN}^{\text{t2i}})$ for each text $T_i$. In the traditional setting of contrastive learning, the ground truth is derived from the annotation of the dataset, which means that the labeled image-text pairs in the dataset are considered to be semantically consistent, while the arbitrary unlabeled image and text in the dataset are considered to have no semantic associations. Therefore, a one-hot label $\mathbf{y_{i}}=
(y_{i1},...,y_{iN})$ is introduced, where $y_{ii}$ equals 1 for positive pairs and 0 for others. Then the infoNCE loss can be expressed as

\begin{equation}
\label{equ:loss_itc_i2t}
    \mathcal{L}_{\text{itc}}^{\text{i2t}}=\frac{1}{N}\sum_{i=1}^{N}\mathcal{H}\left(\mathbf{y_i}, Q_{i}^{\text{i2t}}\right),
\end{equation}

\begin{equation}
\label{equ:loss_itc_t2i}
    \mathcal{L}_{\text{itc}}^{\text{t2i}}=\frac{1}{N}\sum_{i=1}^{N}\mathcal{H}\left(\mathbf{y_i}, Q_{i}^{\text{t2i}}\right).
\end{equation}
where $\mathcal{H}(.,.)$ denotes the cross-entropy operation, and the final loss is denoted as $\mathcal{L}_{\text{itc}}=\left(\mathcal{L}_{\text{itc}}^{\text{i2t}}+\mathcal{L}_{\text{itc}}^{\text{t2i}}\right)/2$. 
We use $\mathcal{L}_{\text{original}}$ to represent the original loss function of the ITR model, which is equal to $\mathcal{L}_{\text{itc}}$ for most models. This is illustrated in Figure \ref{fig:method_all} with a green circle plus sign.

\subsection{Feature Extraction}

We introduce two uni-modal pre-training models as teacher models to calculate soft-labels for guiding the ITR model. In this work, we select Unicom \cite{anxiang_2023_unicom} for the teacher model of images and Sentence-BERT \cite{reimers2019sentencebert} for the texts. Unicom is currently the state-of-the-art model for image retrieval, while Sentence-BERT is a well-known model that achieves outstanding performance in the tasks of Semantic Textual Similarity (STS). During training, we extract image features from all images in the datasets with Unicom-ViT-B/32 \footnote{\url{https://github.com/deepglint/unicom}} and text features from all available text using MPNet \cite{song2020mpnet} fine-tuned by the authors of Sentence-BERT, i.e. all-mpnet-base-v2 \footnote{\url{https://huggingface.co/sentence-transformers/all-mpnet-base-v2}}. The extraction of image and text features can be done offline, so it does not add complexity to the ITR model during online training. It is worth noting that the choice of teacher models for images and texts can be flexible and can be replaced with any available models.

\subsection{Cross-modal Soft-label Alignment}

In practice, there may be a potential semantic association between the unpaired image and text in the same batch, but not labeled in the dataset. We call this situation ``the inter-modal matching missing problem'', which leads to semantically matched images and text being incorrectly pushed away during training. To address the problem, we propose the Cross-modal Soft-label Alignment (CSA) method (Figure \ref{fig:method_all}). Specifically, we calculate the cosine similarity between $\hat{I}_i$ and $\hat{I}_j$ based on the features obtained from the teacher model and denoted it as $r_{ij}^{\text{i2i}}$. Then the similarity between $\hat{I}_i$ and $\hat{I}_j$ is performed within-batch normalization to obtain $P_{ij}^{\text{i2i}}$, the probability estimate that these two images are semantically consistent from the teacher network:
\begin{equation}
\label{equ:distribution_P}
    P_{ij}^{\text{i2i}}={\frac{\exp\left(r_{ij}^{\text{i2i}}\right)}{\sum_{j=1}^{N}\exp\left(r_{ij}^{\text{i2i}}\right)}}.
\end{equation}
Finally we denote the probability distribution $(P_{i1}^{\text{i2i}},...,P_{iN}^{\text{i2i}})$ as $P_{i}^{\text{i2i}}$. Similarly, we calculate the similarity between $T_i$ and $T_j$, denoted as $r_{ij}^{\text{t2t}}$ and obtain $P_{i}^{\text{t2t}}$. During training, we regard $P_i^{\text{i2i}}$ as the target distribution to guide the learnable distribution $Q_{i}^{\text{i2t}}$ for image-to-text alignment using KL divergence. Similarly, we use $P_i^{\text{t2t}}$ to guide the learnable distribution $Q_{i}^{\text{t2i}}$ for text-to-image alignment using KL divergence. Finally, the loss function for CSA is denoted as $\mathcal{L}_{\text{CSA}}$, which can be written as

\begin{equation}
\label{equ:loss_CSA}
\begin{split}
    \mathcal{L}_{\text{CSA}} &= \left(\mathcal{L}_{\text{CSA}}^{\text{i2t}}+\mathcal{L}_{\text{CSA}}^{\text{t2i}}\right) / 2 \\
    &= \left(D_{KL}(P_i^{\text{i2i}} \parallel Q_i^{\text{i2t}})+D_{KL}(P_i^{\text{t2t}} \parallel Q_i^{\text{t2i}})\right) / 2.
\end{split}
\end{equation}

\begin{table*}[ht]
\small
\setlength{\tabcolsep}{1.8mm}
\centering
\begin{tabular}{lcccccccccccccc} 
\toprule
\multirow{3}{*}{Model} & \multicolumn{7}{c}{MSCOCO (5K Test Set)}                                                                              & \multicolumn{7}{c}{Flickr30K (1K Test Set)}                                                                            \\
                       & \multicolumn{3}{c}{Image-to-Text}             & \multicolumn{3}{c}{Text-to-Image}             & \multirow{2}{*}{RSUM} & \multicolumn{3}{c}{Image-to-Text}             & \multicolumn{3}{c}{Text-to-Image}             & \multirow{2}{*}{RSUM}  \\
                       & R@1           & R@5           & R@10          & R@1           & R@5           & R@10          &                       & R@1           & R@5           & R@10          & R@1           & R@5           & R@10          &                        \\ 
\midrule
\multicolumn{15}{l}{\textit{~~~ Faster-RCNN, ResNet-101, without pre-training}}                                                                                                                                                                                                     \\
SCAN                   & 50.4          & 82.2          & 90.0          & 38.6          & 69.3          & 80.4          & 410.9                 & 67.4          & 90.3          & 95.8          & 48.6          & 77.7          & 85.2          & 465.0                  \\
VSE$\infty$              & 56.6          & 83.6          & 91.4          & 39.3          & 69.9          & 81.1          & 421.9                 & 76.5          & 94.2          & 97.7          & 56.4          & 83.4          & 89.9          & 498.1                  \\
VSRN++                 & 54.7          & 82.9          & 90.9          & 42.0          & 72.2          & 82.7          & 425.4                 & 79.2          & 94.6          & 97.5          & 60.6          & 85.6          & 91.4          & 508.9                  \\
NAAF                   & 58.9          & 85.2          & 92.0          & 42.5          & 70.9          & 81.4          & 430.9                 & \textbf{81.9} & \textbf{96.1} & 98.3          & 61.0          & 85.3          & 90.6          & 513.2                  \\
SGR$\dagger$                    & 57.3          & 83.2          & 90.6          & 40.5          & 69.6          & 80.3          & 421.5                 & 76.6          & 93.7          & 96.6          & 56.1          & 80.9          & 87.0          & 490.9                  \\
~ + \textbf{CUSA}               & 57.4          & 84.5          & 92.0          & 40.9          & 71.2          & 81.9          & 427.9                 & 79.3          & 94.9          & 97.5          & 58.4          & 84.2          & 89.5          & 503.7                  \\
SAF$\dagger$                    & 55.5          & 83.8          & 91.8          & 40.1          & 69.7          & 80.4          & 421.3                 & 75.6          & 92.7          & 96.9          & 56.5          & 82.0          & 88.4          & 492.1                  \\
~ + \textbf{CUSA}               & 55.6          & 84.7          & 92.3          & 40.8          & 71.7          & 82.4          & 427.5                 & 77.8          & 95.0          & 98.0          & 58.5          & 83.9          & 90.3          & 503.5                  \\
SGRAF$\dagger$                  & 58.8          & 84.8          & 92.1          & 41.6          & 70.9          & 81.5          & 429.7                 & 78.4          & 94.6          & 97.5          & 58.2          & 83.0          & 89.1          & 500.8                  \\
~ + \textbf{CUSA}               & \textbf{59.8} & \textbf{86.1} & \textbf{93.3} & \textbf{43.3} & \textbf{73.2} & \textbf{83.6} & \textbf{439.2}        & 81.4          & 95.6          & \textbf{98.5} & \textbf{61.0} & \textbf{86.1} & \textbf{91.5} & \textbf{514.1}         \\ 
\midrule
\multicolumn{15}{l}{\textit{~~~ Dual-Encoder, pre-training}}                                                                                                                                                                                                            \\
CLIP$_\text{ViT-B/32}$            & 56.3          & 81.7          & 89.4          & 42.8          & 71.2          & 81.1          & 422.6                 & 78.7          & \textbf{95.4} & \textbf{98.0} & 66.3          & 88.6          & 93.1          & 520.0                  \\
~ + \textbf{CUSA}       & \textbf{57.3} & \textbf{83.1} & \textbf{90.3} & \textbf{44.2} & \textbf{72.7} & \textbf{82.1} & \textbf{429.7}        & \textbf{82.1} & 95.3          & 97.9          & \textbf{67.5} & \textbf{89.6} & \textbf{93.9} & \textbf{526.3}         \\
CLIP$_\text{ViT-L/14}\ddagger$            & 67.1          & 89.4          & 94.7          & 51.6          & 79.1          & 87.7          & 469.6                 & 87.3          & 99.0          & 99.5          & 76.4          & 94.8          & 97.4          & 554.5                  \\
~ + \textbf{CUSA}       & \textbf{67.9} & \textbf{90.3} & \textbf{94.7} & \textbf{52.4} & \textbf{79.8} & \textbf{88.1} & \textbf{473.1}        & \textbf{90.8} & \textbf{99.1} & \textbf{99.7} & \textbf{77.4} & \textbf{95.5} & \textbf{97.7} & \textbf{560.2}         \\ 
\midrule
\multicolumn{15}{l}{\textit{~~~ Dual Encoder + Fusion encoder reranking, pre-training}}                                                                                                                                                                                 \\
BLIP$_\text{base}$            & 81.9          & 95.4          & 97.8          & 64.3          & 85.7          & 91.5          & 516.6                 & 97.3          & 99.9          & 100.0         & 87.3          & 97.6          & 98.9          & 581.0                  \\
OmniVL                 & 82.1          & 95.9          & 98.1          & 64.8          & 86.1          & 91.6          & 518.6                 & 97.3          & 99.9          & 100.0         & 87.9          & 97.8          & 99.1          & 582.0                  \\
X2VLM$_\text{base}$           & \textbf{83.5}          & 96.3          & 98.5          & 66.2          & 87.1          & 92.2          & 523.8                 & 98.5          & 100.0         & 100.0         & 90.4          & 98.2          & 99.3          & 586.4                  \\
~ + \textbf{CUSA}               & 83.3             & \textbf{96.6}             & \textbf{98.5}             & \textbf{67.1}             & \textbf{87.6}             & \textbf{92.7}             & \textbf{525.8}                     & \textbf{98.5}             & \textbf{100.0}             & \textbf{100.0}             & \textbf{91.3}             & \textbf{98.8}             & \textbf{99.5}             & \textbf{588.1}                      \\
\textcolor{gray}{X2VLM$_\text{large}$}          & \textcolor{gray}{84.4} & \textcolor{gray}{96.5} & \textcolor{gray}{98.5}                 & \textcolor{gray}{67.7} & \textcolor{gray}{87.5} & \textcolor{gray}{92.5}                 & \textcolor{gray}{527.1}                 & \textcolor{gray}{98.8} & \textcolor{gray}{100.0} & \textcolor{gray}{100.0}               & \textcolor{gray}{91.8} & \textcolor{gray}{98.6} & \textcolor{gray}{99.5}                 & \textcolor{gray}{588.7}                  \\
\bottomrule
\end{tabular}
\caption{Experimental results of image-text retrieval on MSCOCO and Flickr30K. $\dagger$ denotes the improved results by the author compared to the original paper, while $\ddagger$ represents the CLIP$_{\text{ViT-L/14@336px}}$ model.}
\label{tab:itr_mscoco_flickr}
\end{table*}
\subsection{Uni-modal Soft-label Alignment}
\begin{figure}[ht]
\centering
\includegraphics[width=\linewidth]{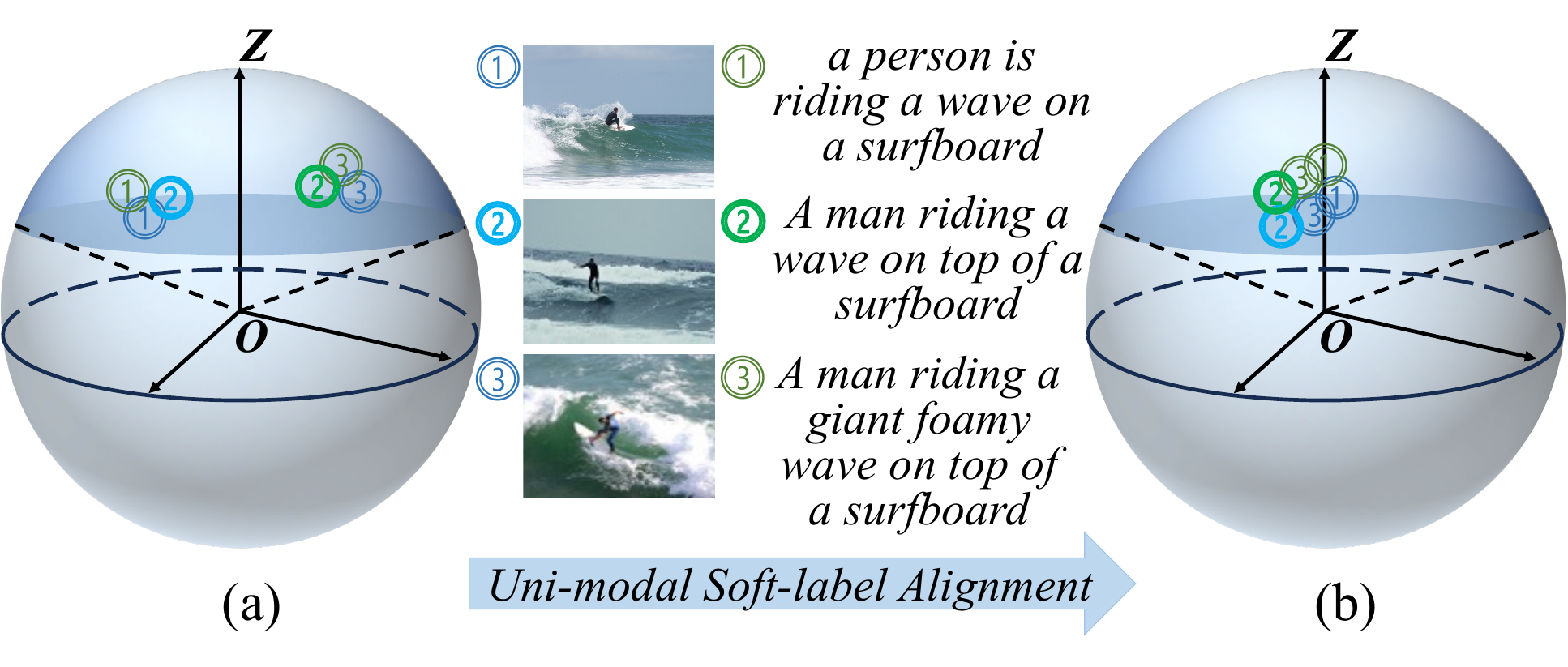}
\caption{(a) Models ignoring intra-modal alignment tend to obtain feature distributions on the hypersphere; (b) After adding the USA term, the model tends to obtain feature distributions on the hypersphere.}
\label{fig:two_ball}
\end{figure}

Although many works in ITR have achieved impressive results, they neglect the uni-modal alignment. In this work, we call this situation ``the intra-modal semantic loss problem'', which may affect the model's generalization performance on unseen data. As shown in Figure \ref{fig:two_ball}, we consider the scenario where the image-text pair \ding{172} and \ding{174} are the samples in the training set, while the pair \ding{173} is an unseen sample during training. For most models in ITR, it is likely to encounter the situation depicted in Figure \ref{fig:two_ball}(a): the image and text in each pair can be aligned well, but two pairs may be mapped to different regions on the hypersphere since uni-modal alignment is not introduced. Assume that the encoder that completes the training is L-Lipschitz continuity, which means that it maps the elements that are close enough in the sample level to near positions on the hypersphere. Consider the case where the encoder generalizes to sample \ding{173}: image \ding{173} is closer to image \ding{174} at the pixel level and is therefore mapped to its proximity, and text \ding{173} is closer to text \ding{174} at the literal level and is therefore also mapped to its proximity. As a result, it is difficult for image \ding{173} to be recalled by text \ding{173} relying on a specific distance function of the hypersphere, and vice versa. In addition, we mathematically prove that the models that only focus on cross-modal alignment in ITR are deficient in uni-modal capabilities, precisely because the ability to recognize similar input samples of the model is not good enough, which limits the generalization performance of cross-modal retrieval.
\begin{prop}
\label{lemma:cm-insuff-um}
Cross-modal alignment alone is not sufficient for optimal recognition of similar samples.\footnote{Please refer to Appendix A for the proof: \url{https://github.com/lerogo/aaai24_itr_cusa}}
\end{prop}

Motivated by Proposition~\ref{lemma:cm-insuff-um}, we propose the Uni-modal, Soft-label Alignment (USA) method to enhance the model's ability to recognize the similarity between uni-modal samples, thereby improving performance in the unseen data. As illustrated in Figure \ref{fig:method_all}, we first obtain $P_i^{\text{i2i}}$ and $P_i^{\text{t2t}}$ from the teacher model, respectively. followed by extracting the representation $hat{I}$ for image $I$ and $\hat{T}$ for text $T$ from the outputs of the ITR model. Note that each representation is then passed through an additional projector, which is a Full Connectivity (FC) layer in practice. Similar to the Cross-modal Soft-label Alignment method, we express the cosine similarity of $\hat{I}_i$ and $\hat{I}_j$ as $s_{ij}^{\text{i2i}}$, while that of $\hat{T}_i$ and $\hat{T}_j$ as $s_{ij}^{\text{t2t}}$. Then the similarity between $\hat{I}_i$ and $\hat{I}_j$ is performed within-batch normalization to obtain $Q_{ij}^{\text{i2i}}$, the probability estimate that these two images are semantically consistent from student network:
\begin{equation}
\label{equ:distribution_Q_i2i}
    Q_{ij}^{\text{i2i}}=\frac{\exp\left(s_{ij}^{\text{i2i}} / \tau \right)}{\sum_{j=1}^{N}\exp\left(s_{ij}^{\text{i2i}}/ \tau\right)}.
\end{equation}
Finally we denote the probability distribution $(Q_{i1}^{\text{i2i}},...,Q_{iN}^{\text{i2i}})$ as $Q_{i}^{\text{i2i}}$ and use the similar step to obtain $Q_{i}^{\text{t2t}}$. During training, we employ KL divergence to guide the alignment loss of the uni-modal logits using $P_i^{\text{i2i}}$ and $P_i^{\text{t2t}}$, respectively. This alignment loss facilitates the alignment between uni-modal samples: During training, we regard $P_i^{\text{i2i}}$ as the target distribution to guide the learnable distribution $Q_{i}^{\text{i2i}}$ for image-to-text alignment using KL divergence. Similarly, we use $P_i^{\text{t2t}}$ to guide the learnable distribution $Q_{i}^{\text{t2t}}$ for text-to-image alignment. Finally, the loss function for USA is denoted as $\mathcal{L}_{\text{USA}}$, which can be written as

\begin{equation}
\label{equ:loss_USA}
\begin{split}
    \mathcal{L}_{\text{USA}} &= \left(\mathcal{L}_{\text{USA}}^{\text{i2i}}+\mathcal{L}_{\text{USA}}^{\text{t2t}}\right) / 2 \\
    &= \left(D_{KL}(P_i^{\text{i2i}} \parallel Q_i^{\text{i2i}})+D_{KL}(P_i^{\text{t2t}} \parallel Q_i^{\text{t2t}})\right) / 2.
\end{split}
\end{equation}
The USA method brings similar samples closer together in a uni-modal manner while pushing dissimilar samples apart. After such an intra-modal alignment operation, we can make the model more biased towards the situation in Figure \ref{fig:two_ball} (b). Through this uni-modal constraint, we hope that the ITR model can achieve better results in the unseen data.

\subsection{Training Objective}

We use the above two losses, CSA and USA together, to adjust the original loss of the ITR model, so the overall loss function is expressed as:
\begin{equation}
\label{equ:all_loss}
    \mathcal{L}_{\text{CUSA}} = \mathcal{L}_{\text{original}}+\alpha \cdot \mathcal{L}_{\text{CSA}}+ \beta \cdot \mathcal{L}_{\text{USA}}.
\end{equation}
where $\alpha$ and $\beta$ is the loss weight, which ranged from 0.1 to 1.0. Our method is plug-and-play and does not affect the original architecture of the ITR model. When applying our method, we only need to add one FC layer at the image and text ends respectively to implement the USA method, and the rest of the model structure does not need any changes. Therefore, it can be easily extended to existing ITR models.

\section{Experiments}
\subsection{Experiment Setup}
\subsubsection{Datasets}
\begin{table}[t]
\small
\setlength{\tabcolsep}{1.0mm}
\centering
\begin{tabular}{lcccccc} 
\toprule
\multirow{2}{*}{Model}                          & \multicolumn{3}{c}{Image-to-Text}             & \multicolumn{3}{c}{Text-to-Image}              \\
                          & mAP@R         & R-P           & R@1           & mAP@R         & R-P           & R@1            \\ 
\midrule
\multicolumn{7}{l}{\textit{~~~ Faster-RCNN, ResNet-101, without pre-training}}                                                                  \\
SGR$\dagger^{1}$                      & 26.8          & 38.7          & 70.3          & 42.2          & 51.2          & 83.6           \\
~ + \textbf{CUSA}         & 28.0          & 40.0          & 72.4          & 44.0          & 53.0          & 83.4           \\
SAF$\dagger^{1}$                      & 26.6          & 38.5          & 69.6          & 43.1          & 52.0          & 83.8           \\
~ + \textbf{CUSA}         & 27.4          & 39.8          & 71.4          & 44.4          & 53.6          & 84.6           \\
SGRAF$\dagger^{1}$                    & 28.1          & 39.8          & 72.3          & 43.7          & 52.5          & 84.4           \\
~ + \textbf{CUSA}         & \textbf{29.5} & \textbf{41.4} & \textbf{74.5} & \textbf{46.4} & \textbf{55.1} & \textbf{85.7}  \\ 
\midrule
\multicolumn{7}{l}{\textit{~~~ Dual-Encoder, pre-training}}                                                                         \\
CLIP$_{\text{ViT-B/32}}$      & 28.5          & 39.4          & \textbf{72.5} & 41.7          & 50.8          & 83.0           \\
~ + \textbf{CUSA} & \textbf{29.6} & \textbf{40.7} & 72.0          & \textbf{45.2} & \textbf{53.6} & \textbf{85.7}  \\
CLIP$_\text{ViT-L/14@336px}$      & 32.8          & 43.4          & 79.7          & 45.5          & 54.2          & 87.2           \\
~ + \textbf{CUSA} & \textbf{33.6} & \textbf{44.1} & \textbf{80.9} & \textbf{47.6} & \textbf{55.8} & \textbf{88.2}  \\ 
\midrule
\multicolumn{7}{l}{\textit{~~~ Dual Encoder + Fusion encoder reranking, pre-training}}                                              \\
X2VLM$_\text{base}\dagger^{2}$                    & 36.6          & 45.2          & 89.7          & 43.8          & 51.2          & 93.5           \\
~ + \textbf{CUSA}         & \textbf{37.6}             & \textbf{46.5}             & \textbf{89.8}             & \textbf{48.4}             & \textbf{55.9}             & \textbf{94.1}              \\
\bottomrule
\end{tabular}

\caption{Experimental results of image-text retrieval on ECCV Caption. $\dagger^{1}$ denotes the results of reproducing the method, and $\dagger^{2}$ denotes the results from the checkpoint provided by the author.}
\label{tab:itr_eccv}
\end{table}

To evaluate the ability of the ITR model in image-text retrieval and judging similar samples, we evaluated our method on several datasets for both cross-modal and uni-modal tasks. For image-text retrieval, we evaluate our approach on three datasets: Flickr30K \cite{young2014image}, MSCOCO \cite{lin2014microsoft}, and ECCV Caption \cite{chun2022eccv_caption}. For image retrieval experiments, our evaluation is conducted on the test sets of four widely used datasets: CUB \cite{cub_welinder2010caltech}, SOP \cite{sop_oh2016deep}, In-Shop \cite{inshop_liu2016deepfashion}, and iNaturalist \cite{inat_van2018oisin}. In terms of semantic textual similarity, we evaluate our approach on seven STS tasks: STS 2012–2016\cite{sts12_agirre2012semeval, sts13_agirre2013sem, sts14_agirre2014semeval,sts15_agirre2015semeval,sts16_agirre2016semeval}, STS Benchmark \cite{stsb_cer2017semeval} and SICK-Relatedness \cite{sick_marelli2014semeval}. Similar to image retrieval, we solely use the test sets of the STS datasets for our semantic textual similarity evaluation.

\subsubsection{Evaluation Metrics}
In the evaluation of the performance on the Flickr30K and MSCOCO datasets, we utilize the R@K(recall at K, $K\in\{1,5,10\}$) metric, representing the proportion of queries where the ground truth is ranked within the top K. Additionally, to comprehensively assess the image-text retrieval performance, we summarize all the recall values as RSUM. Inspired by \cite{chun2022eccv_caption}, we also utilize the R-P and mAP@R metrics on the ECCV Caption dataset to evaluate the ability of the model to recall incorrect negatives. Following the work of \cite{anxiang_2023_unicom}, we adopt the R@1 metric as the standard for evaluating performance across all image retrieval datasets. In the case of semantic textual similarity, we leverage the SentEval \cite{conneau2018senteval} toolkit to compute Spearman's correlation, which serves as a reliable measure of the semantic textual similarity performance of the model.

\subsubsection{Implementation Details}
To validate the improved performance of our approach in cross-modal and uni-modal tasks, we executed a series of experiments involving three models: SGRAF \cite{sgraf_diao2021similarity}, CLIP$_\text{ViT-B/32, ViT-L/14@336}$ \cite{clip_radford2021learning}, and X2VLM$_\text{base}$ \cite{x2vlm_zeng2023x2vlm}. These models are all impressive models for image-text retrieval, with SGRAF being an open-source, non-pretrained SOTA model, CLIP being a popular pre-trained dual-encoder model, and X2VLM being an advanced pre-trained model with dual encoders and a fusion encoder. All CLIP model reports are based on fine-tuned results using InfoNCE.

\subsection{Main Results}
\subsubsection{Results on MSCOCO and Flickr30K}
Table \ref{tab:itr_mscoco_flickr} shows the results of our comparison on various types of ITR models. It shows that our method can achieve improvement on all models and has achieved new SOTA results in both pre-trained and non-pretrained benchmarks. On the MSCOCO 5K test set, our method increased the RSUM of SGRAF$_{\text{+CUSA}}$ by 9.5\%, CLIP$_{\text{B/32+CUSA}}$ by 7.1\%, CLIP$_{\text{L/14+CUSA}}$ by 3.5\%, and X2VLM$_{\text{base+CUSA}}$ by 2.0\%. This is a significant progress. On the Flickr30K test set, our method increased the RSUM of SGRAF$_{\text{+CUSA}}$ by 13.3\%, CLIP$_{\text{B/32+CUSA}}$ by 6.3\%, CLIP$_{\text{L/14+CUSA}}$ by 5.7\%, and X2VLM$_{\text{base+CUSA}}$ by 1.7\%. CUSA consistently performs excellently, proving its effectiveness and robustness.
\begin{table}[t]
\small
\setlength{\tabcolsep}{1.0mm}
\centering
\begin{tabular}{lcccccc} 
\toprule
Model            & CUB  & SOP  & In-Shop & INaturalist & Avg.                    \\
\midrule
\multicolumn{6}{l}{\textit{~~~ Faster-RCNN, ResNet-101, without pre-training}}          \\
SGR$\dagger^{1}$             & 31.1 & 51.9 & 19.5    & 33.7        & 34.1                    \\
~ + \textbf{CUSA}         & \textbf{34.6} & \textbf{60.7} & \textbf{31.6}    & \textbf{41.9}        & \textbf{42.2}                    \\
SAF$\dagger^{1}$             & 34.1 & 52.8 & 20.3    & 37.0        & 36.0                    \\
~ + \textbf{CUSA}         & \textbf{39.9} & \textbf{59.6} & \textbf{32.2}    & \textbf{44.6}        & \textbf{44.1}                    \\
\midrule
\multicolumn{6}{l}{\textit{~~~ Dual-Encoder, pre-training}}                               \\
CLIP$_{\text{ViT-B/32}}$      & 41.5 & 51.8 & 28.1    & 41.3        & 40.7                    \\
~ + \textbf{CUSA} & \textbf{49.6} & \textbf{56.5} & \textbf{34.1}    & \textbf{45.6}        & \textbf{46.5}                    \\
CLIP$_{\text{ViT-L/14@336px}}$      & 58.3 & 61.1 & 46.9    & 63.5        & 57.4                    \\
~ + \textbf{CUSA} & \textbf{67.2} & \textbf{63.0} & \textbf{48.2}    & \textbf{68.7}        & \textbf{61.8}                    \\
\midrule
\multicolumn{6}{l}{\textit{~~~ Dual Encoder + Fusion encoder reranking, pre-training}}    \\
X2VLM$_\text{base}\dagger^{2}$           & 53.6 & 64.2 & 52.6    & 59.3        & 57.4                    \\
~ + \textbf{CUSA}         & \textbf{58.9}    & \textbf{67.0}    & \textbf{54.2}       & \textbf{62.2}           & \textbf{60.6}                       \\
\bottomrule
\end{tabular}
\caption{Performance of image retrieval on 4 datasets.}
\label{tab:i2i_result}
\end{table}
\begin{table}[t]
\small
\setlength{\tabcolsep}{1.0mm}
\centering
\begin{tabular}{lcccccc} 
\toprule
Model            & STS12-16$_{\text{Avg.}}$ & STS-B & SICK-R & Avg.$\ddagger$                       \\ 
\midrule
\multicolumn{5}{l}{\textit{~~~ Faster-RCNN, ResNet-101, without pre-training}}                        \\
SGR$\dagger^{1}$             & 51.8           & 58.1  & 62.7   & 54.3                        \\
~ + \textbf{CUSA}         & \textbf{55.9}           & \textbf{65.2}  & \textbf{64.9}   & \textbf{58.5}                        \\
SAF$\dagger^{1}$             & 53.9           & 64.5  & 63.5   & 56.8                        \\
~ + \textbf{CUSA}         & \textbf{54.8}           & \textbf{66.3}  & \textbf{64.5}   & \textbf{57.8}                        \\ 
\midrule
\multicolumn{5}{l}{\textit{~~~ Dual-Encoder, pre-training}}                               \\
CLIP$_{\text{ViT-B/32}}$      & 67.4           & 76.2  & 72.9   & 69.4                        \\
~ + \textbf{CUSA} & \textbf{71.6}           & \textbf{78.3}  & \textbf{75.8}   & \textbf{73.2}                        \\
CLIP$_{\text{ViT-L/14@336px}}$      & 69.8           & 78.6  & \textbf{75.5}   & 71.9                        \\
~ + \textbf{CUSA} & \textbf{73.4}           & \textbf{79.9}  & 74.9   & \textbf{74.5}                        \\ 
\midrule
\multicolumn{5}{l}{\textit{~~~ Dual Encoder + Fusion encoder reranking, pre-training}}    \\
X2VLM$_\text{base}\dagger^{2}$           & 26.6           & 22.3  & 50.4   & 29.4                        \\
~ + \textbf{CUSA}         & \textbf{46.8}              & \textbf{47.9}     & \textbf{76.2}      & \textbf{51.2}                           \\
\bottomrule
\end{tabular}
\caption{Sentence embedding performance on STS tasks. $\ddagger$ represents the average result of 7 STS datasets.}
\label{tab:t2t_result}
\end{table}
\begin{table}[ht]
\small
\setlength{\tabcolsep}{0.8mm}
\centering
\begin{tabular}{lccccc} 
\toprule
\multirow{2}{*}{Model} & \multicolumn{3}{c}{Cross-modal}                     & \multicolumn{2}{c}{Uni-modal}  \\
                       & RSUM$\ddagger^{1}$ & RSUM$\ddagger^{2}$ & Avg.$\ddagger^{3}$ & Avg.$_{\text{IR}}$ & Avg.$_{\text{STS}}$           \\ 
\midrule
CLIP$_{\text{ViT-B/32}}$            & 422.6        & 520.0           & 52.6               & 40.7     & 69.4                \\
\text{~ + CSA}             & 427.7        & 525.4           & 54.5               & 40.4     & 66.5                \\
\text{~ + USA}             & 425.7        & 523.4           & 53.6               & 47.8     & 73.1                \\
\text{~ + CUSA}            & 429.7        & 526.3           & 54.5               & 46.5     & 73.2                \\
\bottomrule
\end{tabular}
\caption{Ablation study on two types of tasks with four different settings. $\ddagger^{1}$, $\ddagger^{2}$, and $\ddagger^{3}$ represent results on MSCOCO, Flickr30K, and ECCV Caption datasets respectively.}
\label{tab:ablation_clip}
\end{table}

\subsubsection{Results on ECCV Caption}
We also conducted a fair experiment on the ECCV Caption dataset. As shown in Table \ref{tab:itr_eccv}, with the help of our CUSA method, there is an average improvement of 1.0\% in image-to-text retrieval for 4 models on 3 metrics, while in text-to-image retrieval, the average improvement of all metrics is 2.5\%. The results show that the introduction of our CUSA method can improve the accuracy and recall of false negatives retrieved by ITR models, proving the effectiveness of our method.

\subsection{Additional Results on Uni-Modal Retrieval}
Table \ref{tab:i2i_result} and Table \ref{tab:t2t_result} show the performance of the ITR model in uni-modal tasks. In the image modality, our CUSA method shows an average improvement of 5.9\% compared to models without it. In the text modality, the average improvement is 6.7\%. This strongly indicates that our method can enhance the ability of the model to recognize similar input samples, thus facilitating image-text retrieval. Not only that, the significant improvement in uni-modal tasks brings a certain universal retrieval ability to the ITR model, which can be very useful in low-resource scenarios.
\begin{figure}[ht]
\centering
\includegraphics[width=0.98\linewidth]{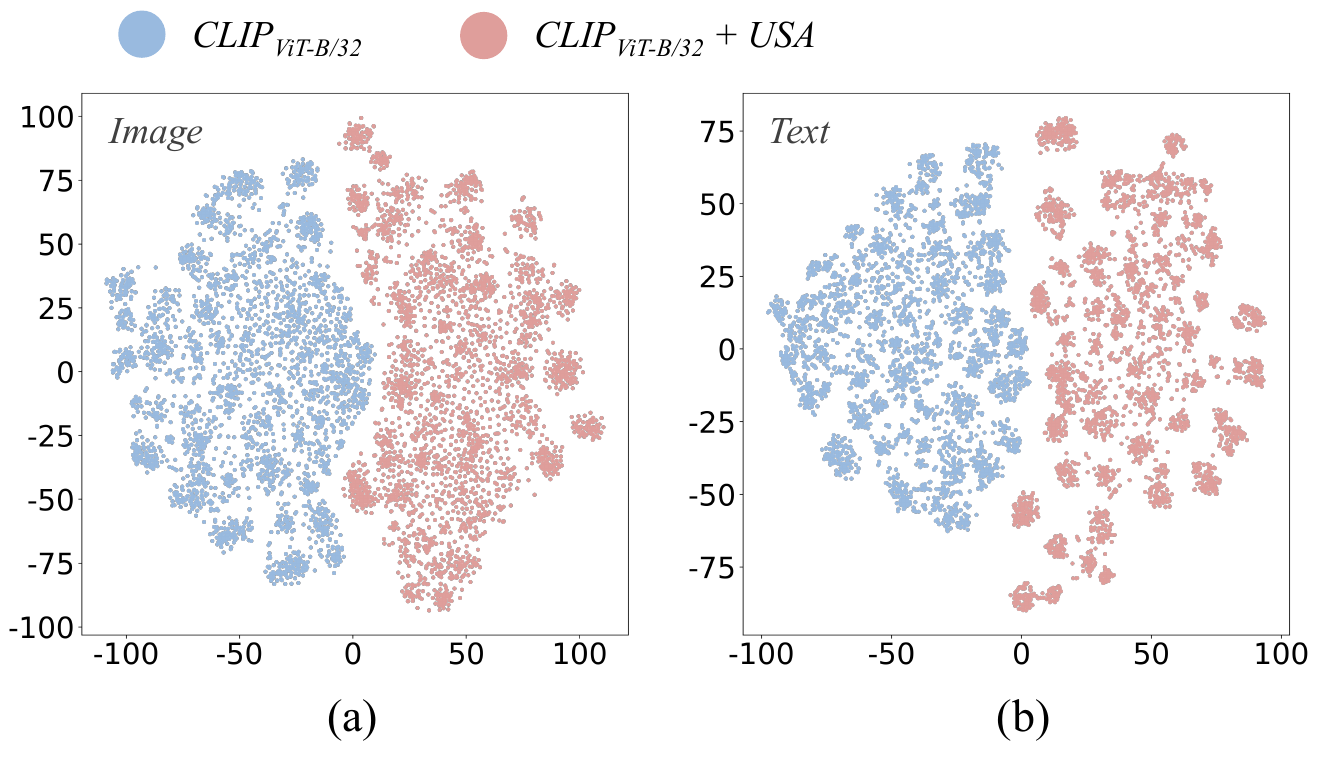}
\caption{Visualization of features generated from 5000 randomly selected image-text pairs from the MSCOCO test set. (a) represents the visualization of image features, while (b) represents the visualization of text features.}
\label{fig:ablation_usa}
\end{figure}
\begin{figure}[ht]
\centering
\includegraphics[width=\linewidth]{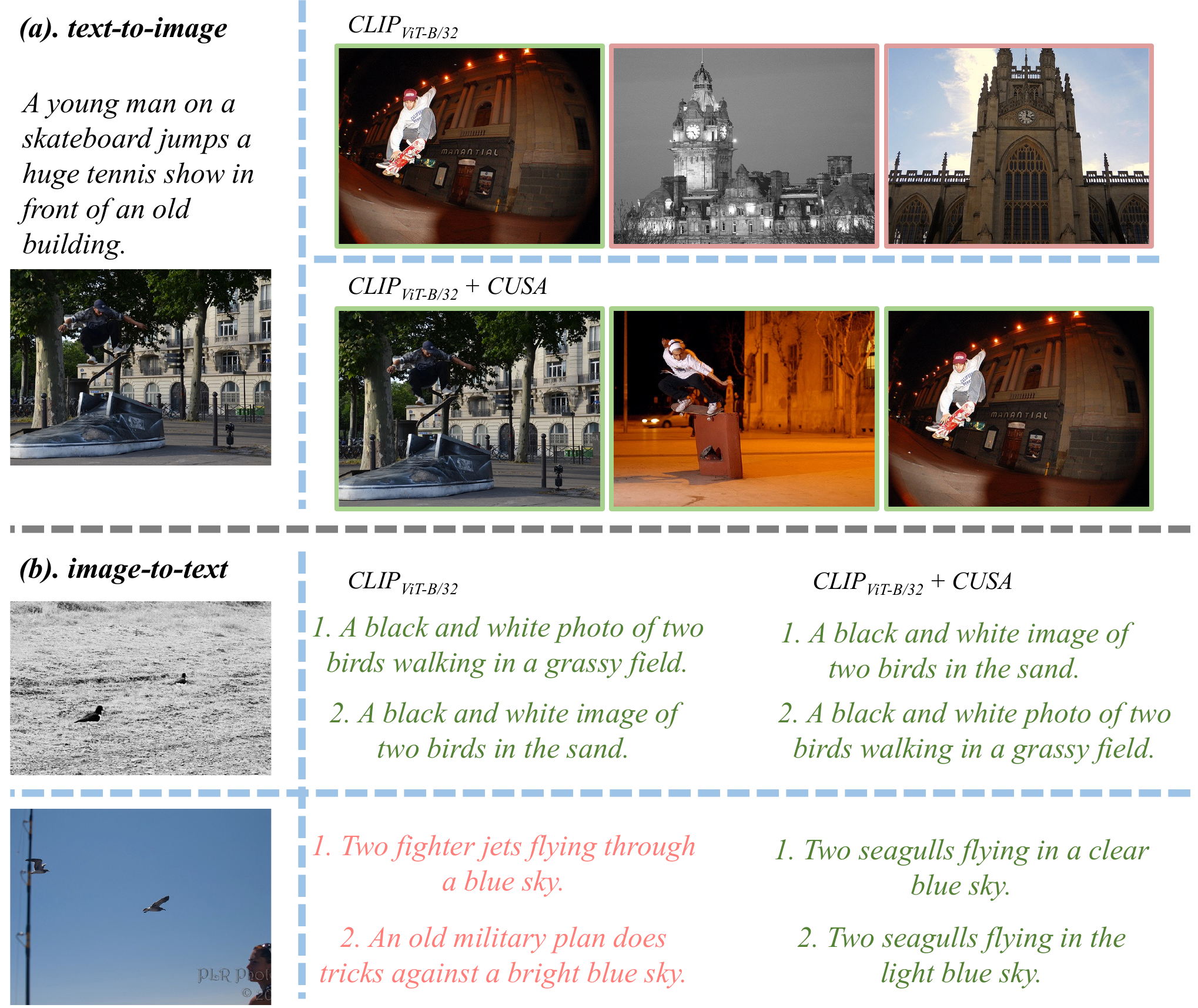}

\caption{Case study: the green texts or boxes represent the same as the ground-truth, while the red ones do not.}
\label{fig:case_study}
\end{figure}
\subsection{Ablation Study}
To evaluate the contributions of CSA and USA, we used the original loss and tested each method on various ITR models. In this analysis, we fine-tuned CLIP$_{\text{ViT-B/32}}$ on the MSCOCO dataset and evaluated its performance on two types of tasks, as shown in Table \ref{tab:ablation_clip}. The results show that removing either CSA or USA leads to a decrease in image-text retrieval performance, with the effect of CSA being more significant. The reason is that CSA aims to overcome false negatives and improve the cross-modal capabilities of the model. Without the USA method for uni-modal alignment, it would significantly impair the ability of ITR models to recognize similar input samples, thereby harming performance.

To demonstrate the impact of the USA method on the features generated by the ITR models, we conducted a controlled experiment on the CLIP$_{\text{ViT-B/32}}$ model both with and without the USA method. We fine-tuned it on MSCOCO, randomly selected 5000 image-text pairs from its test set, and generated features for images and texts separately. Then we used TSNE for dimensionality reduction analysis, and the visualization results are shown in Figure \ref{fig:ablation_usa}. It shows that the features generated with the USA method and those generated without the USA method are completely distinguished, indicating that the USA method has changed the distribution of the original feature (Whether in image modality or text modality). Further observation of the visualization illustration shows that the USA method can better cluster similar samples together and distinguish dissimilar samples, indicating that the USA method is effective in improving the ability of the ITR model to recognize similar input samples.

\subsection{Case Study}
We have presented the comparison results between our method and the original model in Figure \ref{fig:case_study}. As shown in Figure \ref{fig:case_study} (a), we utilized a text query to retrieve the top-3 ranked images, and the results indicate that our method can recall more false negative cases. Furthermore, in Figure \ref{fig:case_study} (b), we employed two similar images to retrieve the top-2 similar texts, and the results demonstrate that our method enables the model to better recognize similar input samples.

\section{Conclusion}
In this paper, we have proposed a novel method for image-text retrieval, called Cross-modal and Uni-modal Soft-label Alignment. Our method leverages a uni-modal pre-training model to provide soft-label supervision signals for the ITR model, and uses two alignment techniques, CSA and USA, to overcome false negatives and enhance similarity recognition between uni-modal samples. Our method is plug-and-play and can be easily applied to existing ITR models without changing their original architectures. We have conducted extensive experiments on various ITR models and datasets and demonstrated that our method can consistently improve the performance of image-text retrieval and achieve new state-of-the-art results. Moreover, our method can also boost the uni-modal retrieval performance of the ITR model, enabling it to achieve universal retrieval.

\newpage

\bibliography{aaai24}

\end{document}